\documentclass[10pt]{article} 
\usepackage[preprint]{tmlr}

\usepackage{amsmath,amsfonts,bm}

\def\eqref#1{equation~\ref{#1}}

\def\1{\bm{1}}

\DeclareMathAlphabet{\mathsfit}{\encodingdefault}{\sfdefault}{m}{sl}
\SetMathAlphabet{\mathsfit}{bold}{\encodingdefault}{\sfdefault}{bx}{n}

\usepackage{hyperref}
\usepackage{url}

\usepackage{microtype}
\usepackage{graphicx}
\usepackage{xcolor}
\usepackage{hyperref}
\usepackage{url}
\usepackage{booktabs}
\usepackage{amsmath}
\usepackage{multirow}
\usepackage{tcolorbox}

\usepackage{lineno}

\definecolor{darkblue}{rgb}{0, 0, 0.5}
\definecolor{revisiongreen}{rgb}{0, 0.6, 0}

\title{Organizing, Orchestrating, and Benchmarking Agent Skills at Ecosystem Scale}

\author{Hao Li\thanks{Equal contribution}\quad%
Chunjiang Mu\footnotemark[1]\quad%
Jianhao Chen\quad%
Siyue Ren\quad%
Zhiyao Cui\quad\\
Yiqun Zhang\quad%
Lei Bai\quad%
Shuyue Hu\thanks{Corresponding author. Email: hushuyue@pjlab.org.cn.}\\
Shanghai Artificial Intelligence Laboratory
}

\def\month{MM}  
\def\year{YYYY} 
\def\openreview{\url{https://openreview.net/forum?id=XXXX}} 

\begin{document}

\maketitle

\begin{abstract}
The rapid proliferation of Claude agent skills has raised the central question of how to effectively leverage, manage, and scale the agent skill ecosystem. 
In this paper, we propose \emph{AgentSkillOS}, the \textbf{first} principled framework for skill selection, orchestration, and ecosystem-level management.
\emph{AgentSkillOS} comprises two stages: (i) \textbf{Manage Skills}, which organizes skills into a capability tree via node-level recursive categorization for efficient discovery; and (ii) \textbf{Solve Tasks}, which retrieves, orchestrates, and executes multiple skills through DAG-based pipelines. 
To evaluate the agent's ability to invoke skills, we construct a benchmark of 30 artifact-rich tasks across five categories: data computation, document creation, motion video, visual design, and web interaction.
We assess the quality of task outputs using LLM-based pairwise evaluation, and the results are aggregated via a Bradley--Terry model to produce unified quality scores.
Experiments across three skill ecosystem scales ($200$ to $200\text{K}$ skills) show that tree-based retrieval effectively approximates oracle skill selection, and that DAG-based orchestration substantially outperforms native flat invocation even when given the identical skill set.
Our findings confirm that structured composition is the key to unlocking skill potential. Our GitHub repository is available at: \href{https://github.com/ynulihao/AgentSkillOS}{\textcolor{blue}{https://github.com/ynulihao/AgentSkillOS}}.
\end{abstract}

\section{Introduction}\label{sec:intro}
The agent skill ecosystem is rapidly expanding~\citep{ling2026agent}. Introduced by Claude in October 2025,
a skill is a markdown file that defines a structured folder comprising declarative instructions, executable scripts, and auxiliary resources~\citep{anthropic_agent_skills_overview}. By supporting dynamic loading and execution at runtime, skills enable a large language model (LLM) to acquire domain-specific knowledge and extend its operational capabilities~\citep{xu2026agent}.
Skills have been adopted by multiple model service providers and integrated into platforms such as Coze, enabling cross-model skill reuse and execution. As of late Feb 2026, more than 280,000 skills are publicly available,\footnote{https://skillsmp.com/} and the overwhelming majority is developed and maintained by decentralized, third-party contributors.
At this scale and degree of decentralization, a critical question emerges:
\textit{how can the agent skill ecosystem be effectively leveraged, managed, and scaled?}

This question is of both practical importance and research interest.
From a user perspective, the sheer number of available skills makes it difficult to obtain a global view of the ecosystem. Users often lack visibility into what skills exist, what capabilities they expose, and how they differ from or overlap with one another, making targeted skill selection for specific tasks challenging. From a platform provider perspective, a rapidly expanding, largely third-party ecosystem creates major governance challenges, especially around quality and reliability~\citep{liu2026malicious, schmotz2025agent}.
More fundamentally, as the ecosystem scales, skills increasingly become fragmented and isolated contributions; without explicit mechanisms for composition and coordination, many skills remain underused, and the ecosystem fails to deliver its key value: orchestrating multiple skills to solve tasks beyond any single skill~\citep{li2026single}.



In this paper, we introduce \emph{AgentSkillOS}, the first principled framework for skill selection, orchestration, and ecosystem-level management.
It consists of two stages: \textbf{Manage Skills }and \textbf{Solve Tasks}. In the Manage Skills stage, we construct a \textbf{capability tree} that organizes skills into a hierarchy based on their capabilities.
Specifically, starting from the root node, \emph{AgentSkillOS} organizes skills into a node-level hierarchy by \textbf{recursively partitioning} each node's skill set into category groups, splitting any category node that exceeds a per-node capacity threshold until all nodes conform to the limit.
We note that the construction of a capability tree is essential. It surfaces non-obvious yet functionally relevant skills, thereby enabling broader and more creative skill discovery than pure semantic retrieval.
In the Solve Tasks stage, \emph{AgentSkillOS} builds on the capability tree to retrieve and orchestrate multiple skills, equipping a task-specific agent to solve tasks beyond any single skill. This stage comprises three steps: (i) \textbf{Task-driven Skill Retrieval}, which explores the capability tree to retrieve a broad set of relevant skills and then prunes them via filtering, deduplication, and relevance ranking; (ii) \textbf{DAG-based Skill Orchestration}, which decomposes the task into subtasks and composes the selected skills into Directed-Acyclic-Graph (DAG) plans, offering three different strategies for orchestration; and (iii) \textbf{Multi-skill Task Execution}, which executes the selected DAG, automatically managing execution order, dependencies, and data flow across steps.

To systematically evaluate \emph{AgentSkillOS}, we construct a benchmark of 30 tasks across five categories: data computation, document creation, motion video, visual design, and web interaction. All tasks are constructed by human experts who first curate high-quality skills from public marketplaces and GitHub repositories, then craft task descriptions and deliverable requirements based on these skills and real-world user needs, either from a single skill or by cross-composing multiple skills.
The benchmark features three key properties tailored for skill evaluation: (i)~\textbf{Multi-format Creative Tasks}: each task requires delivering a complete, end-user-facing artifact, going beyond code generation or question answering to test real-world delivery quality; (ii)~\textbf{Pairwise Comparison with Position-bias Mitigation}: we evaluate via LLM-based pairwise judging conducted in both orderings to counteract position bias, and aggregate preferences via a Bradley--Terry model~\citep{bradley1952rank} for fine-grained ranking; and (iii)~\textbf{High Skill-discriminability}: by grounding tasks in specific skills, task quality depends critically on whether the agent correctly selects and invokes the relevant skills, clearly surfacing differences in skill configuration.

We evaluate different configurations across three skill ecosystem sizes ($200$, $1\text{K}$, $200\text{K}$).
Results show that our \emph{AgentSkillOS} consistently achieves the highest scores across all ecosystem sizes, substantially outperforming both the vanilla Claude Code skill invocation and the skill-free baseline.
Notably, our ablation study reveals that DAG-based skill orchestration is a critical contributor to performance: even when provided with the identical oracle skill set, the vanilla Claude Code agent, which invokes skills in a flat, unstructured manner, performs significantly worse than \emph{AgentSkillOS} with DAG orchestration.
This demonstrates that structured skill composition, not merely skill availability, is the key to unlocking the full potential of the skill ecosystem.

The main contributions of this paper are summarized as follows:
\begin{itemize}
    \item We propose \emph{AgentSkillOS}, the first principled framework comprising capability-tree-based skill organization and DAG-based multi-skill orchestration, for efficiently leveraging, managing, and scaling agent skill ecosystems.
    \item We construct a benchmark of 30 artifact-rich tasks across five categories with an LLM-based pairwise evaluation protocol aggregated via a Bradley--Terry model for fine-grained quality assessment.
    \item Through comprehensive experiments across three ecosystem scales ($200$ to $200\text{K}$ skills), we demonstrate that tree-based retrieval effectively approximates oracle skill selection, and that DAG-based orchestration substantially outperforms native flat skill invocation even when given the oracle skill set.
\end{itemize}

\section{Method}

\begin{figure}[t]
  \centering
  \includegraphics[width=1.0\linewidth]{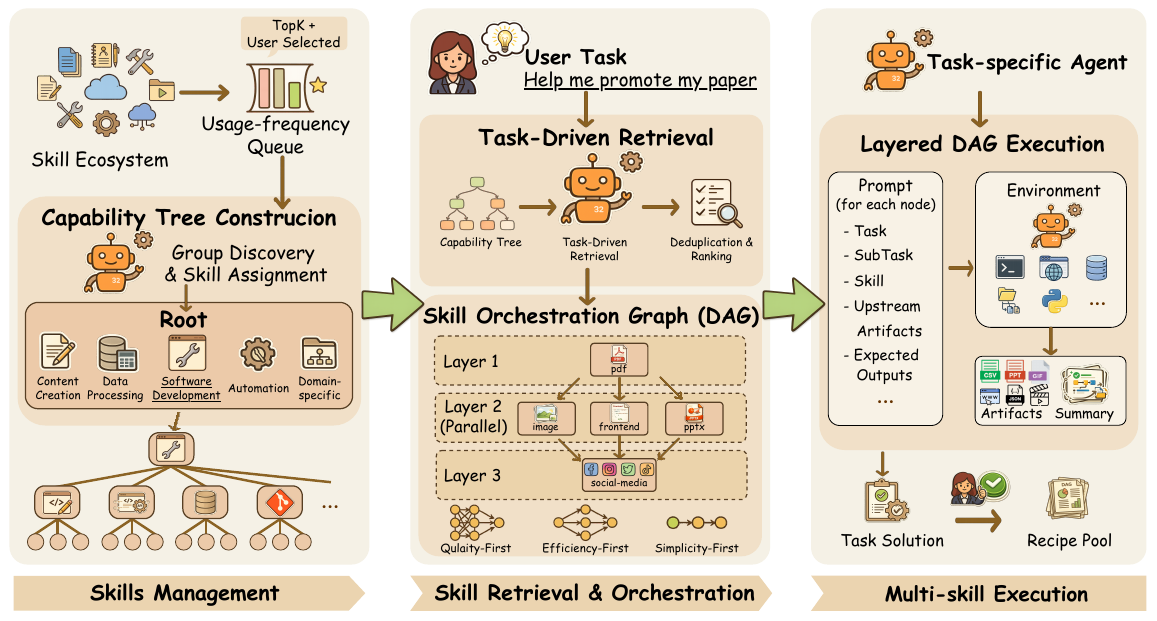}
  \caption{\emph{AgentSkillOS} is a principled framework for efficient skill retrieval, orchestration, and ecosystem-level management to solve user-specified tasks.}
  \label{fig:framework}
\end{figure}

In this section, we present \emph{AgentSkillOS}, the first principled framework for efficiently selecting, orchestrating, and managing skills from a large-scale ecosystem to solve a user-specified task.  
\emph{AgentSkillOS} comprises two stages: (i) \textbf{Manage Skills}, which organizes a large-scale skill ecosystem into a capability tree to enable efficient use of available skills prior to task execution; 
and (ii) \textbf{Solve Tasks}, which automatically selects, orchestrates, and executes multiple skills during task execution.
Fig.~\ref{fig:framework} shows the overall workflow of \emph{AgentSkillOS}.

\subsection{Manage Skills: Capability Tree Construction}
To enable efficient retrieval of required skills from a large-scale skill ecosystem, we organize the skill ecosystem using a tree structure.
We refer to this tree as the \textit{capability tree}, which is constructed and updated offline, independent of the task execution process. 
Formally, we denote skill ecosystem to be managed by $S$, and the capability tree by $T$. 
Each node $n\in T$ corresponds to a skill partition and is assigned a specific subset of skills $\mathcal{S}_n\subseteq \mathcal{S}_T$, where $\mathcal{S}_T$ is the set of all skills used to construct the capability tree $T$.
Therefore, for the root node $r$, it holds that $\mathcal{S}_r=\mathcal{S}_T$.
For node $n$ with children $\mathrm{ch}(n)$, the node's children $c\in \mathrm{ch}(n)$ form a partition of $\mathcal{S}_n$ which satisfies $\mathcal{S}_n=\bigcup_{c\in \mathrm{ch}(n)}\mathcal{S}_c$ and $\mathcal{S}_c\cap \mathcal{S}_{c'}=\emptyset$ for any $c\neq c'$. 
From high-level nodes to low-level nodes of this capability tree, the skill number of each level of nodes decreases approximately exponentially. Ultimately, each leaf node in the capability tree corresponds to a skill partition containing exactly one skill.
This structure of our capability tree supports coarse-to-fine, level-by-level localization: agents can first identify relevant top-level capability domains and then search within the corresponding node partitions, avoiding unguided retrieval over $\mathcal{S}_T$. It is worth noting that $\mathcal{S}_T$ is not necessarily equal to the entire skill ecosystem, which we denote as $\mathcal{S}$.
In the following subsections, we first describe how the capability tree is constructed and updated through node-level recursive categorization over $\mathcal{S}_T$. We further introduce a usage-frequency queue to select $\mathcal{S}_T$ from $\mathcal{S}$ to control the tree size and reduce the inclusion of low-quality skills when $\mathcal{S}$ becomes excessively large.

\subsubsection{Node-Level Recursive Categorization}
Prior to capability tree construction, \emph{AgentSkillOS} scans $\mathcal{S}_T$ and extracts name and description of each skill as the basis for subsequent partition.
Then, starting from the root node, \emph{AgentSkillOS} adopts a breadth-first strategy to generate child nodes for each node, thereby building the capability tree $T$ level by level. 
The child nodes are generated by partitioning the skill sets of the current node, which consists of two steps: Group Discovery and Skill Assignment. 
Given the current node $n$ and its assigned skill set $\mathcal{S}_n$, during Group Discovery, \emph{AgentSkillOS} employs an LLM to generate a specific number of category groups for $\mathcal{S}_n$, including the name and description of each category. 
The target number of groups is guided by a branching factor $B$. 
In Skill Assignment, \emph{AgentSkillOS} employs an LLM to assign each skill in $\mathcal{S}_n$ to one of the generated categorie groups. 
By splitting children generation into two steps, \emph{AgentSkillOS} greatly reduces the probability of missing skills during tree construction. 

\emph{AgentSkillOS} performs additional handling for three special cases. (i) If the current node is the root node, to ensure stable and reliable classification, five fixed category groups are manually specified in the Group Discovery: content creation, data processing, software development, automation, and domain-specific. (ii) If the number of skills in any generated category after Skill Assignment is equal to 1, that category and all its skills will be merged into the most relevant target category, and the name and description of the target category will be updated. If the number of skills in a category is below the per-node capacity threshold $C$ but greater than or equal to 2, all skills in that category are converted into leaf nodes, and no further child nodes of them will be generated. By ensuring that each node is assigned a sufficient number of skills, \emph{AgentSkillOS} reduces the depth of the capability tree and accelerating skill retrieval. (iii) If some skills are not assigned to any category because of LLM hallucination or other bugs, these skills will be reassigned. If they still have no category after the second Skill Assignment, they will be assigned to the category with the largest number of skills.
Ultimately, each node in the capability tree corresponds to a category of skills, while each leaf node represents an individual skill.

The capability tree will be updated if ${S}$ changes. The newly introduced skills are incorporated into the capability tree as follows.
Starting from the root node, each new skill is progressively assigned to the most appropriate category at each level, forming a path down the tree and ultimately being inserted as a new leaf node at the proper position. After insertion, we perform a bottom-up update along the constructed path to refine the category names and descriptions of the affected nodes. Overall, our tree construction and update method achieves a balance between skill classification accuracy and retrieval efficiency.

\subsubsection{Usage-Frequency Queue for Skill Selection for Tree-construction}
We control the depth of the tree by selecting only the skills that satisfy specific criteria from $\mathcal{S}$  to construct the capability tree, when the number of skills in $\mathcal{S}$ is above the threshold $K$.
To select $\mathcal{S}_T$, \emph{AgentSkillOS} maintains a usage-frequency queue $Q$ that ranks $s\in \mathcal{S}$ by a frequency score $f(s)$,  which we set as the install count of skill $s$ on the marketplace\footnote{https://skills.sh/}.
Then, $\mathcal{S}_T$ is selected to construct the tree by $\mathcal{S}_T=\mathrm{TopK}(Q,K)\cup \mathcal{S}^{\mathrm{user}}$, where $\mathcal{S}^{\mathrm{user}}$ denotes the skills manually selected by the user from the $\mathcal{S}$.
This design enables users to conveniently incorporate their own custom and private skills.
We place the remaining skills $\mathcal{S} \setminus \mathcal{S}_T$ into a dormant index that supports semantic suggestions. 
Specifically, we build a vector index over dormant-skill metadata (name and description) and retrieve suggestions by embedding similarity.
If a suggested dormant skill proves useful during task solving, \emph{AgentSkillOS} allows users to add it to $\mathcal{S}^{\mathrm{user}}$ and trigger the tree update.


\subsection{Solve Tasks: a Task-Specific Agent with Multiple Skills}
Building on the capability tree of skills, \emph{AgentSkillOS} retrieves and orchestrates multiple skills to equip a task-specific agent, enabling it to solve tasks beyond the reach of any single skill and improving overall skill utilization. 
The key to constructing a task-specific agent lies in generating and executing a skill orchestration graph for the given task, formulated as a Directed Acyclic Graph (DAG).
We denote it by $G = (V,E)$, where each node $v\in V$ denotes a skill in the capability tree and is selected to solve the task, and each directed edge $(u, v) \in E$ indicates that skill $v$ depends on the outputs or effects of skill $u$.
In practice, different nodes typically correspond to distinct stages of task execution. 
By exposing relevant skills at specific stages of the task, \emph{AgentSkillOS} enables the agent to invoke skills more accurately.
The task-solving process is organized into three stages:
(i) Task-driven Skill Retrieval, which retrieves the skills that serve as nodes in the skill orchestration graph; (ii) DAG-based Skill Orchestration, which constructs the DAG by organizing the retrieved skills; and (iii) Multi-skill Task Execution, which completes the task based on the DAG.
We will introduce three stages in the following sections.
Incidentally, \emph{AgentSkillOS} introduces a skill orchestration reuse mechanism based on task description vector similarity to improve the operational efficiency. 
\textbf{For similar user tasks, this mechanism allows the system to directly reuse the skill orchestration plans in the recipe pool with user consent, thereby skipping stage 1 and stage 2.}


\subsubsection{Task-driven Skill Retrieval}
\emph{AgentSkillOS} selects skills for a user-specified task in two stages: \emph{retrieval} and \emph{pruning}.
Leveraging the capability tree, \emph{AgentSkillOS} prompts the LLM to traverse the hierarchy and select relevant category nodes layer by layer based on the user's task, ultimately treating all reached leaf nodes as candidate skills.
In this process, the LLM is explicitly encouraged to include any skill that could help achieve the user's goal.
For skills not included in the tree (if any), \emph{AgentSkillOS} uses a vector similarity search to retrieve the skills most relevant to the task and adds them to the candidate skill set.
Compared with keyword search or purely embedding-based retrieval, this tree-guided approach enables the LLM to leverage reasoning and creativity to surface complementary, non-obvious skills. 
In the \emph{pruning} stage, \emph{AgentSkillOS} instructs the LLM to deduplicate and rank the retrieved skills according to their relevance to the user's task, discarding those that are clearly irrelevant or redundant.  
Subsequently, \emph{AgentSkillOS} produces a compact shortlist of the top \(M\) ranked skills.
These skills are then presented to the user, who may additionally select skills, thereby determining the final node set \(V\) for constructing the skill orchestration graph $G$.

\subsubsection{DAG-based Skill Orchestration}
Once \(V\) is determined, \emph{AgentSkillOS} generates alternative skill-orchestration plans for the user to select from, based on different pre-defined orchestration strategies. 
The strategies define different structural patterns for organizing skills into an executable graph, reflecting trade-offs among latency, risk, and solution diversity. 
In this paper, we define three such strategies: (i) Quality-First, which maximizes the output quality of every deliverable by fully leveraging each skill's strengths and adding preparation or refinement stages where they improve results; (ii) Efficiency-First, which maximizes parallelism by reducing sequential hops and structuring the DAG so that independent sub-tasks run concurrently whenever possible, and (iii) Simplicity-First, which produces a minimal DAG in which every node is essential---no node can be removed without compromising task completion.

During skill-orchestration generation, \emph{AgentSkillOS} provides the LLM with the user task $t$, the selected skill set $V$, and orchestration strategies to decompose the task into sub-tasks, each of which can be handled by a corresponding skill to improve its performance.
For each sub-task, it specifies two aspects: (i) its dependencies on other sub-tasks (i.e., its inputs rely on outputs produced by other sub-tasks) and (ii) its objective and expected outputs. 
Based on this information, \emph{AgentSkillOS} then organizes the sub-tasks and their corresponding skills into three skill orchestration graphs with different strategies, where dependencies among nodes define the directed edges.
Formally, let $\ell(v)$ denote the layer index for any node $v\in V$, and each dependency satisfies $(u,v)\in E \Rightarrow \ell(u)<\ell(v)$.

\subsubsection{Multi-skill Task Execution}
Once an orchestration with specific strategies is selected by the user, the task-specific agent equipped with the selected skills executes the nodes according to the DAG's layered dependencies:
nodes within the same layer have no dependencies and can be executed in parallel, while those on different layers must be executed sequentially.
For each node, \emph{AgentSkillOS} constructs an execution prompt that restates the user task, specifies the skill to invoke, and clarifies the sub-task. It also lists upstream artifacts (i.e., files produced by prior nodes) with brief usage hints, states the expected outputs, and explains how downstream nodes will consume them. 
The agent then executes the specified skill, saves all generated files as artifacts, and produces a structured summary of execution status and outputs to support inspection.
Specifically, the skill orchestration graph for each task is also stored for reuse in future similar tasks.


\begin{figure}[ht]
  \centering
  \includegraphics[width=0.9\linewidth]{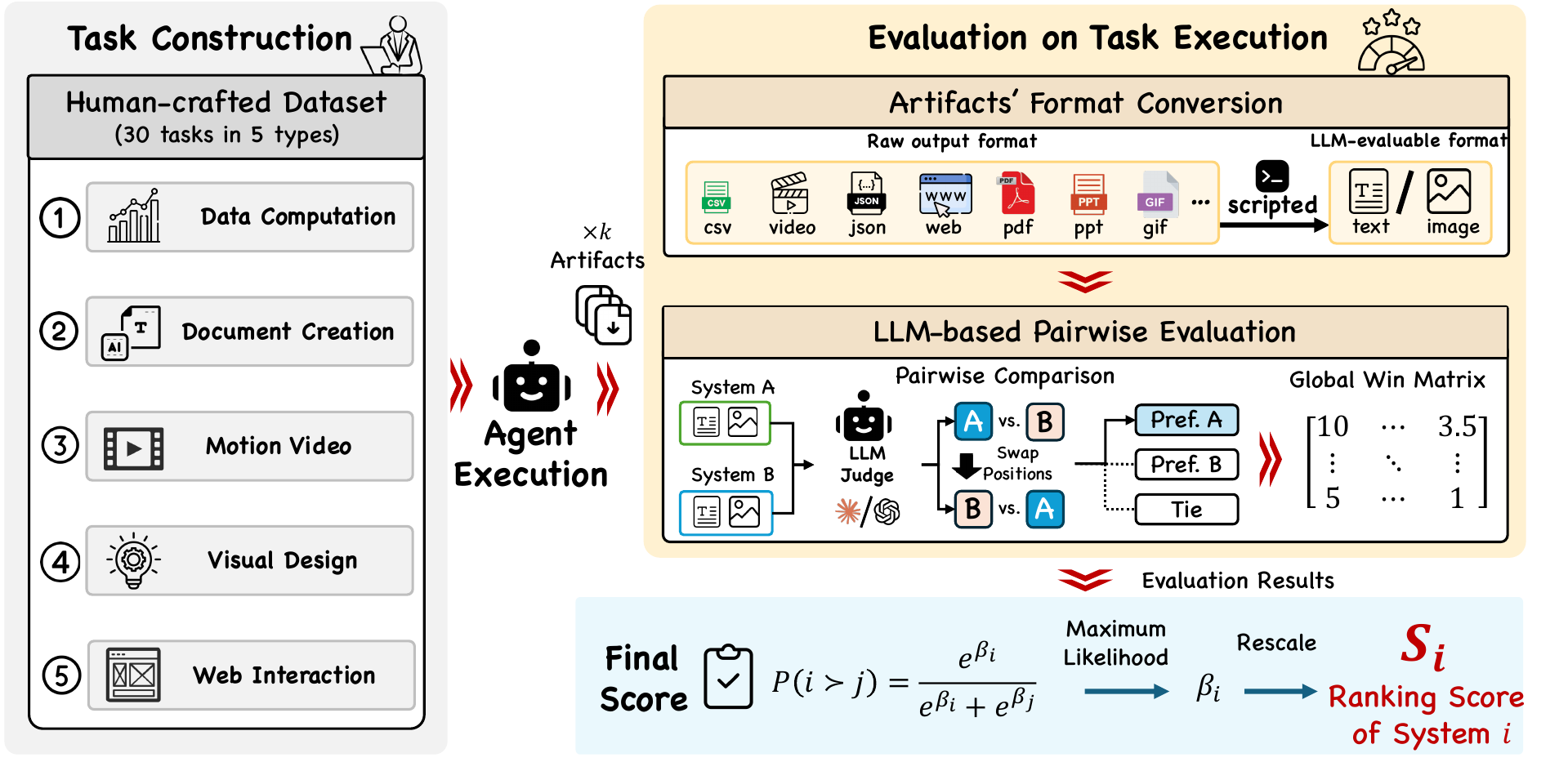}
  \caption{Overview of our benchmark framework. The benchmark consists of a human-crafted dataset of 30 tasks spanning five categories: Data Computation, Document Creation, Motion Video, Visual Design, and Web Interaction. For evaluation, artifacts produced by agents in diverse raw formats (csv, video, json, web, pdf, gif) are first converted into LLM-evaluable formats (text and image) via scripted pipelines. An LLM judge then performs pairwise comparisons between systems with position swapping to mitigate order bias, and the results are aggregated into a global win matrix. Finally, a Bradley--Terry model is fitted via maximum likelihood to obtain strength parameters $\beta_i$, which are rescaled to produce continuous ranking scores $S_i$ for fine-grained differentiation of agent performance.} 
  \label{fig:benchmark}
\end{figure}

\section{Benchmark}
We construct a benchmark of 30 tasks to evaluate agents' ability to discover and invoke skills for generating diverse types of artifacts (see Fig.~\ref{fig:benchmark}).
Our benchmark features three key design principles:
\textbf{(i)~Multi-format Creative Tasks.} Tasks go beyond code generation or question answering: they require producing polished, end-user-facing artifacts in diverse formats (PDF, PPTX, DOCX, HTML pages, videos, and generated images), where layout, design, and aesthetics all matter.
\textbf{(ii)~Pairwise Comparison with Position-bias Mitigation.} Rather than assigning absolute scores, the evaluation asks an LLM judge to select the better output from two systems. Each comparison is conducted in both orderings; consistent preferences are accepted and conflicting ones are recorded as ties, directly counteracting position bias.
\textbf{(iii)~High Skill-discriminability.} Task quality depends critically on whether the agent correctly selects and invokes the corresponding skill---an agent with the right skill produces high-quality output, while one lacking it fails or yields drastically degraded results---ensuring that evaluation clearly surfaces differences in skill configuration.

\subsection{Task}
\begin{figure}[t]
  \centering
  \includegraphics[width=1.0\linewidth]{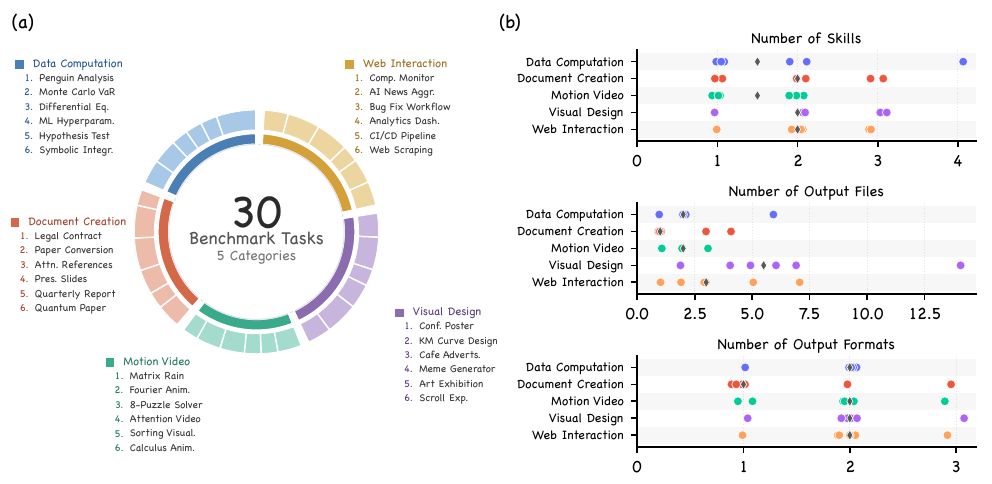}
  \caption{Overview of benchmark tasks. (a)~All 30 tasks organized into five categories: Data Computation, Document Creation, Motion Video, Visual Design, and Web Interaction, with six tasks per category. Each task is listed with its abbreviated name. (b)~Task complexity distributions across categories, measured by three dimensions: the number of skills required to complete the task, the number of output files the task expects, and the number of distinct output formats involved.}
  \label{fig:task_overview}
\end{figure}

The tasks in our benchmark span five categories: \textbf{data computation}, \textbf{document creation}, \textbf{motion video}, \textbf{visual design}, and \textbf{web interaction}.
Data computation tasks involve numerical analysis and statistical modeling; document creation tasks require generating professionally formatted documents and presentations; motion video tasks focus on producing educational animations explaining mathematical and technical concepts; visual design tasks center on designing static visual artifacts; and web interaction tasks encompass building interactive web pages or automated web data collection.
Each task requires delivering a complete, end-user-facing artifact, closely resembling the real-world workflows of human users. 
All tasks are constructed by human experts through the following procedure. We first manually select high-quality skills from public skill marketplaces and GitHub repositories. Based on the scenarios these skills target and the likely needs of real-world users, we then craft task descriptions and specify the required deliverables, either from a single skill or by cross-composing multiple skills.
Figure~\ref{fig:task_overview} (a) lists all 30 tasks organized by category, and (b) shows the complexity profile of each category in terms of the number of skills, output files, and output formats required per task.

\subsection{Evaluation}
To evaluate task artifacts, we adopt pairwise comparative evaluation as the primary metric. Given the outputs of two systems for the same task, an LLM judge examines all artifacts from both sides and determines which system produced the better result, considering dimensions such as correctness, completeness, quality, and aesthetics. To support multi-modal comparison, artifacts in non-text formats are converted into representations consumable by the LLM judge: documents and slides are rendered as page images, HTML pages are captured as full-page screenshots, videos are represented by uniformly sampled frames accompanied by metadata (duration, resolution, frame rate), and images are resized to a standardized resolution. Text files are included verbatim up to a length limit.
To mitigate position bias inherent in LLM-based judging, each comparison is conducted in both orderings. The two judgments are then consolidated into a single verdict: if both orderings agree, the consistent preference is adopted; if exactly one ordering produces an error, the valid judgment is used; if the two orderings yield conflicting preferences (i.e., each favors a different system), the outcome is recorded as a tie.
When ranking $N$ systems, we perform all $\binom{N}{2}$ pairwise comparisons across every shared task. The pairwise outcomes are aggregated into a win matrix $W \in \mathbb{R}^{N \times N}$, where $W_{ij}$ records the number of times system $i$ is preferred over system $j$. Ties contribute $0.5$ to both $W_{ij}$ and $W_{ji}$. We then fit a Bradley-Terry model~\citep{bradley1952rank} to derive a ranking from the win matrix.
The Bradley-Terry model assigns each system $i$ a latent strength parameter $\beta_i \in \mathbb{R}$, and models the probability that system $i$ is preferred over system $j$ as:
\begin{equation}
P(i \succ j) = \frac{e^{\beta_i}}{e^{\beta_i} + e^{\beta_j}}
\label{eq:bt}
\end{equation}
We fit the model via maximum likelihood using the MM algorithm~\citep{hunter2004mm} with Laplace smoothing ($\alpha=1$), yielding centered strength parameters $\{\beta_i\}$ with zero mean. The resulting strength parameters are linearly rescaled to $[0, 100]$ to obtain the final ranking score for each system:
\begin{equation}
S_i = \frac{\beta_i - \beta_{\min}}{\beta_{\max} - \beta_{\min}} \times 100
\label{eq:rescale}
\end{equation}
Overall, this benchmark provides a systematic and reproducible evaluation of an agent's ability to deliver high-quality artifacts for complex, real-world tasks. We use this benchmark to evaluate different settings in the following experiments.

\section{Experiments}
\subsection{Experimental Setup}

\paragraph{Skill ecosystem.}
We construct skill ecosystems of three sizes ($|\mathcal{S}|{=}200$, $1\text{K}$, $200\text{K}$) to evaluate scalability. All skills are sourced from a public skill marketplace and GitHub repositories.
The smallest pool ($200$) consists of two manually curated components: (i)~the best-performing skill for each benchmark task, and (ii)~a set of additional high-quality skills manually selected by human experts.
The two larger pools ($1\text{K}$ and $200\text{K}$) extend the $200$-skill base by automatically including skills ranked by install count from the marketplace until the target size is reached.
This construction progresses from a curated, high-quality core to increasingly open and noisy skill ecosystems, allowing us to assess how retrieval and orchestration degrade as ecosystem size and heterogeneity grow.

\paragraph{Compared methods.}
We evaluate eight configurations organized into four groups.
(i)~\textit{\emph{AgentSkillOS} variants} employ capability-tree-based retrieval and DAG-based orchestration, differing only in orchestration strategy: \textbf{Quality-First} maximizes output quality by adding preparation and refinement stages; \textbf{Efficiency-First} maximizes parallelism by reducing sequential hops; and \textbf{Simplicity-First} produces a minimal DAG in which every node is essential.
(ii)~\textit{Oracle orchestration.} \textbf{Quality-First (Oracle)} bypasses tree retrieval and instead uses the benchmark-designated skills (i.e., the ground-truth relevant skills for each task), while retaining the Quality-First DAG orchestration, serving as an upper-bound reference for skill retrieval.
(iii)~\textit{Claude Code Agent SDK with skills but without orchestration.} Three ablations supply skills to Claude Code without DAG-based composition: \textbf{w/ Full Pool} provides the entire skill ecosystem; \textbf{w/ Retrieval} provides only the tree-retrieved skills; and \textbf{w/ Oracle Skills} provides the benchmark-designated skills. In all three cases, the agent invokes skills in a flat, unstructured manner.
(iv)~\textit{Baseline.} \textbf{Vanilla} runs the native Claude Code Agent SDK without access to any skills.

\paragraph{Models.}
For the \emph{AgentSkillOS} variants (groups~i and~ii), capability tree construction, tree-based skill retrieval, and DAG-based skill orchestration use \emph{claude-opus-4.5} as the LLM backbone, while each DAG node is executed by the Claude Code Agent SDK with \emph{claude-sonnet-4.5}.
For the Claude Code Agent SDK variants (groups~iii and~iv), all configurations use the Claude Code Agent SDK with \emph{claude-sonnet-4.5}.
For benchmark evaluation, the LLM-judge pairwise comparison is implemented using the Claude Code Agent SDK with \emph{claude-opus-4.5} as the judge model.

\paragraph{Hyperparameters.}
The branching factor for capability tree construction is set to $B{=}7$ for the $200$ and $1\text{K}$ ecosystems, and $B{=}12$ for the $200\text{K}$ ecosystem; during Group Discovery, each node is partitioned into $[B{-}3,; B{+}2]$ child groups. The per-node capacity threshold is $C = \lfloor 1.5B \rfloor$. For the $200\text{K}$ ecosystem, the usage-frequency queue retains the top $K{=}10{,}000$ skills as the active set $\mathcal{S}_T$ for tree construction; the remaining skills are placed in a dormant index. During skill retrieval, the top $M{=}8$ ranked skills are retained after pruning.

\subsection{Results}

\begin{table*}[t]
\centering
\caption{Per-category W\,/\,T\,/\,L (wins\,/\,ties\,/\,losses) counts from LLM-judge pairwise comparisons against all other methods. Category abbreviations: Data = Data Computation, Document = Document Creation, Video = Motion Video, Visual = visual Design, Web = Web Interaction. Each pair is judged twice with reversed presentation order and merged into a single debiased decision. The last column reports the Bradley--Terry score rescaled to $[0,100]$, where $100$ corresponds to the top-ranked method and $0$ to the bottom-ranked method within each ecosystem size. \textbf{Bold} indicates the best result per column within each ecosystem size.}
\label{tab:main_results}
\small
\setlength{\tabcolsep}{3.5pt}

\resizebox{0.8\textwidth}{!}{%
\begin{tabular}{cl ccccc c}
\toprule
$|\mathcal{S}|$
  & \textbf{Method}
  & \textbf{Data} & \textbf{Document} & \textbf{Video}
  & \textbf{Visual} & \textbf{Web}
  & \textbf{BT} \\
\midrule
\multirow{5}{*}{$200$}
  & Quality-First    & \textbf{18/1/5} & \textbf{12/6/6} & 11/7/6 & \textbf{17/7/0} & \textbf{20/3/1} & \textbf{100.0} \\
  & Efficiency-First & 13/2/9 & 7/10/7 & 6/9/9 & 11/8/5 & 14/3/7 & 58.5 \\
  & Simplicity-First & 10/4/10 & 7/9/8 & 8/7/9 & 12/6/6 & 13/3/8 & 53.6 \\
  \cmidrule(lr){2-8}
  & w/ Full Pool.    & 5/4/15 & 11/8/5 & \textbf{13/3/8} & 3/4/17 & 2/4/18 & 24.3 \\
  & Vanilla & 6/5/13 & 4/5/15 & 5/8/11 & 2/5/17 & 2/5/17 & 0.0 \\
\midrule
\multirow{5}{*}{$1\text{K}$}
  & Quality-First    & \textbf{14/4/6} & \textbf{11/7/6} & 13/2/9 & 9/10/5 & \textbf{20/2/2} & \textbf{100.0} \\
  & Efficiency-First & 10/4/10 & \textbf{11/5/8} & \textbf{15/1/8} & 8/10/6 & 10/6/8 & 76.1 \\
  & Simplicity-First & 9/3/12 & 10/5/9 & 7/3/14 & 9/8/7 & 11/6/7 & 56.3 \\
  \cmidrule(lr){2-8}
  & w/ Full Pool.    & 11/1/12 & 9/5/10 & 14/2/8 & \textbf{10/4/10} & 4/4/16 & 48.1 \\
  & Vanilla & 9/2/13 & 5/6/13 & 6/2/16 & 5/6/13 & 4/4/16 & 0.0 \\
\midrule
\multirow{5}{*}{$200\text{K}$}
  & Quality-First    & \textbf{15/7/2} & \textbf{15/5/4} & 13/3/8 & 9/4/11 & \textbf{18/1/5} & \textbf{100.0} \\
  & Efficiency-First & 11/7/6 & 12/7/5 & \textbf{14/4/6} & 10/5/9 & 16/2/6 & 89.0 \\
  & Simplicity-First & 8/3/13 & 11/5/8 & 8/3/13 & \textbf{12/5/7} & 14/3/7 & 56.0 \\
  \cmidrule(lr){2-8}
  & w/ Full Pool.    & 7/4/13 & 4/5/15 & 12/2/10 & 8/3/13 & 4/6/14 & 17.2 \\
  & Vanilla & 6/5/13 & 3/8/13 & 6/2/16 & 10/5/9 & 0/4/20 & 0.0 \\
\bottomrule
\end{tabular}%
}

\end{table*}

\paragraph{\emph{AgentSkillOS} substantially outperforms both native skill provisioning and the skill-free baseline, and this advantage persists as the skill ecosystem scales.}
As shown in Table~\ref{tab:main_results}, the three \emph{AgentSkillOS} variants (Quality-First, Efficiency-First, and Simplicity-First) consistently achieve the highest Bradley--Terry scores across all ecosystem sizes, substantially outperforming w/ Full Pool, which in turn outperforms the skill-free Vanilla baseline.
Notably, the w/ Full Pool configuration, which supplies the entire skill ecosystem directly to Claude Code, achieves limited Bradley--Terry scores across all ecosystem sizes. This reflects an inherent limitation of the native skill invocation mechanism in Claude Code. As the pool grows, an increasing fraction of skills becomes invisible to the agent, effectively nullifying the benefit of a larger ecosystem.

\begin{figure*}[h]
\centering
\includegraphics[width=1.0\linewidth]{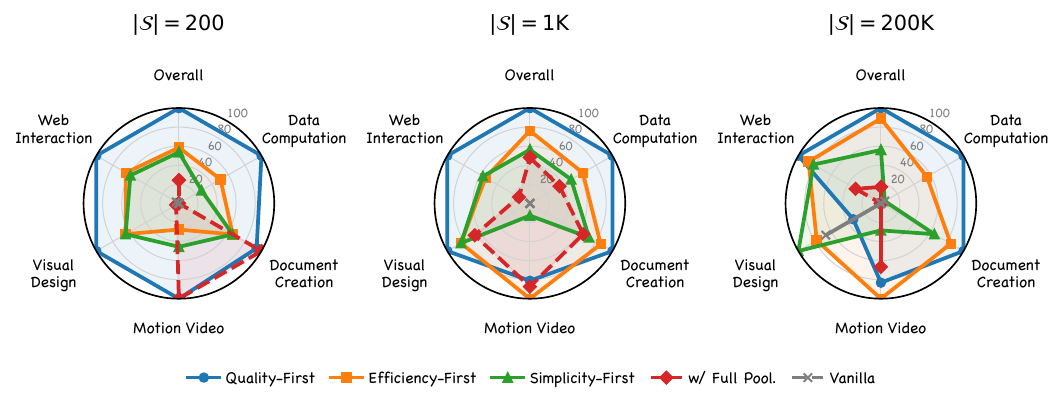}
\caption{Per-category and overall Bradley-Terry scores (rescaled to $[0,100]$) derived from the pairwise comparisons in Table~\ref{tab:main_results}, shown for three skill ecosystem sizes ($|\mathcal{S}|{=}200$, $1\text{K}$, $200\text{K}$). Larger polygon area indicates stronger performance across categories. Three \emph{AgentSkillOS} variants achieve the broadest coverage in all settings, consistent with the top Bradley--Terry scores reported in Table~\ref{tab:main_results}.}
\label{fig:radar}
\end{figure*}

In contrast, the three \emph{AgentSkillOS} variants consistently rank at the top across all ecosystem sizes, demonstrating that structured retrieval via the capability tree and DAG-based orchestration together overcome the scalability bottleneck of flat skill provision.
Fig.~\ref{fig:radar} provides a per-category view by fitting separate Bradley--Terry models to the win matrix of each task category. The resulting radar charts show that the three \emph{AgentSkillOS} variants form large, well-rounded polygons across all ecosystem sizes, whereas w/ Full Pool and Vanilla produce small, irregular shapes with pronounced weaknesses in specific categories, confirming that the advantage holds uniformly across diverse task types.
The limited scores of w/ Full Pool further highlight a fundamental tension between the rapidly expanding skill ecosystem and the agent's inherently constrained ability to discover and invoke skills, underscoring the necessity of principled skill management frameworks such as our \emph{AgentSkillOS}.

\begin{figure*}[h]
\centering
\includegraphics[width=1.0\linewidth]{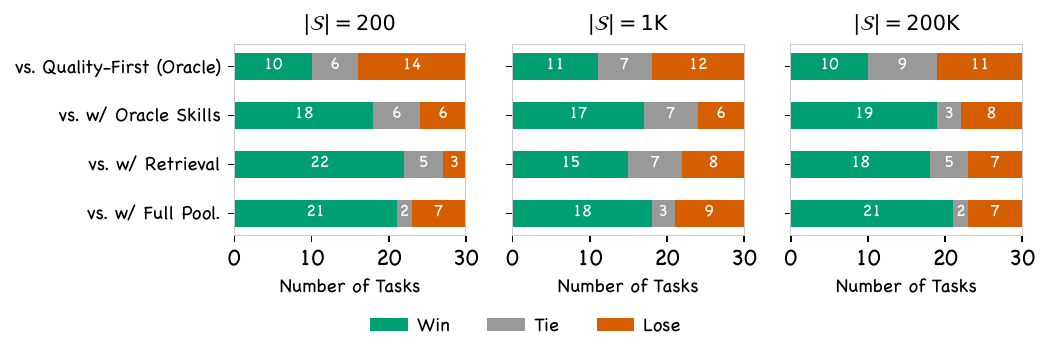}
\caption{Ablation study of \emph{AgentSkillOS} components. Each panel shows pairwise W\,/\,T\,/\,L counts (green\,=\,win, gray\,=\,tie, orange\,=\,lose) of Quality-First against four ablation variants for $|\mathcal{S}|{=}200$, $1\text{K}$, and $200\text{K}$. Removing DAG orchestration (w/ Oracle Skills, w/ Retrieval) or both retrieval and orchestration (w/ Full Pool) consistently degrades performance, confirming that both components are essential. Quality-First closely approaches the oracle upper bound (Quality-First (Oracle)), validating the effectiveness of tree-based skill retrieval.}
\label{fig:ablation}
\end{figure*}

\paragraph{Tree-based retrieval approximates oracle skill selection, and DAG orchestration provides clear additional improvements.}
To disentangle the contributions of \emph{AgentSkillOS}'s two core components, capability-tree-based retrieval and DAG-based orchestration, we conduct an ablation study using Quality-First as the reference and report pairwise W\,/\,T\,/\,L counts against four ablation variants (Fig.~\ref{fig:ablation}).
The results reveal a clear degradation gradient as components are removed.
Quality-First wins most decisively against w/ Full Pool, where neither retrieval nor orchestration is available.
Adding tree-based retrieval alone (w/ Retrieval) reduces the gap but Quality-First still dominates, confirming that retrieval without structured composition is insufficient.
Providing the oracle skill set without orchestration (w/ Oracle Skills) yields further improvement, yet Quality-First retains a clear lead, demonstrating that DAG orchestration contributes substantially even when the perfect skill set is given.
Finally, when compared against Quality-First (Oracle), which combines oracle skills with DAG orchestration, Quality-First shows only a modest deficit, and the gap narrows as the ecosystem grows. This validates that tree-based retrieval effectively approximates oracle skill selection.

\begin{figure*}[h]
\centering
\includegraphics[width=1.0\linewidth]{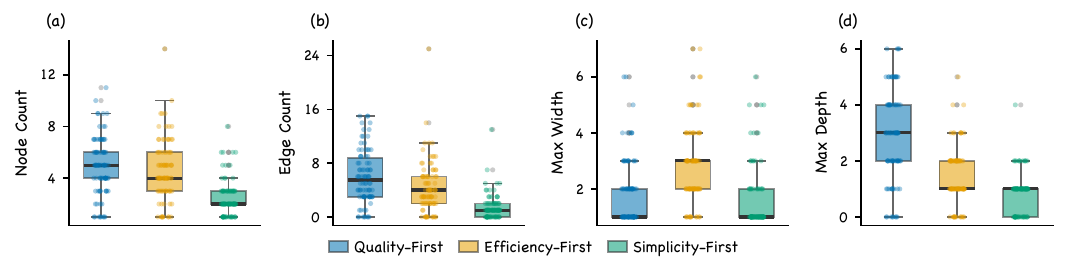}
\caption{Structural properties of DAG-based skill orchestration plans across three strategies, aggregated over all 30 benchmark tasks and three skill ecosystem sizes ($|\mathcal{S}|{=}200$, $1\text{K}$, $200\text{K}$). Each panel reports a different graph-level metric: (a)~node count, (b)~edge count, (c)~max width, and (d)~max depth. Boxes span the interquartile range with individual task results overlaid as jittered points. Quality-First produces the largest and deepest DAGs, reflecting its strategy of adding preparation and refinement stages. Efficiency-First produces graphs of comparable size but with notably greater width and shallower depth, reflecting its preference for maximizing parallel execution over sequential refinement. Simplicity-First yields the most compact graphs, consistent with its minimal-node design objective.}
\label{fig:dag_metrics}
\end{figure*}

\paragraph{Orchestration strategies produce structurally distinct skill graphs that faithfully reflect their design objectives.}                                                                       
As shown in Fig.~\ref{fig:dag_metrics}, the three strategies yield clearly separable distributions over four graph-level metrics: node count, maximum depth (the longest dependency chain, reflecting sequential decomposition), maximum width (the largest number of concurrent nodes at any single layer, reflecting parallelism), and edge count.    
Quality-First consistently produces the largest and deepest graphs with the most edges, indicating that it decomposes tasks into fine-grained, multi-stage pipelines with rich inter-skill dependencies.
Efficiency-First maintains a comparable node count but favors wider, shallower graphs, trading sequential depth for parallel execution. Simplicity-First generates the most compact graphs overall, with the fewest nodes, minimal depth, and sparse connectivity, consistent with its objective of retaining only essential steps. These structural differences confirm that the three strategies do not merely reorder the same set of skills, but shape fundamentally different compositional topologies, providing users with meaningful and interpretable choices when configuring the orchestration pipeline.   

\begin{figure*}[t]
\centering
\includegraphics[width=1.0\linewidth]{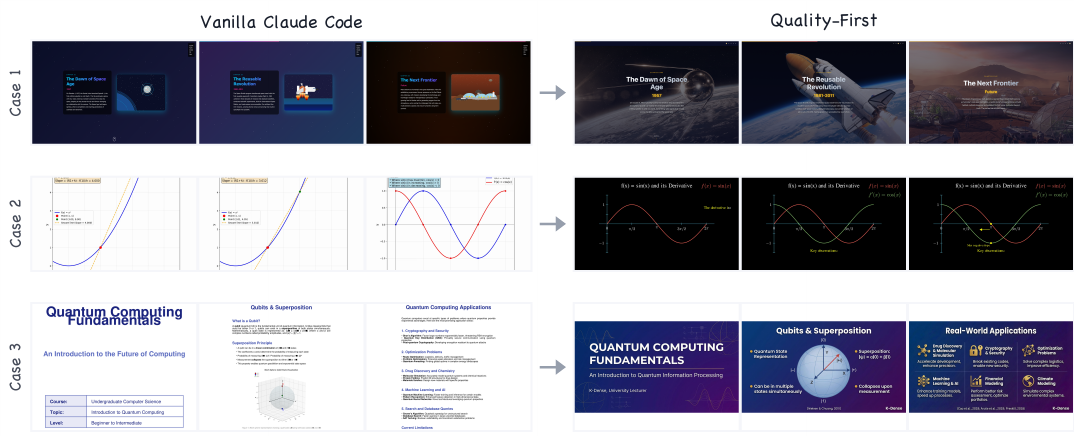}
\caption{Qualitative comparison between the vanilla Claude Code baseline (left) and the Quality-First variant of \emph{AgentSkillOS} (right) on three representative tasks from the $|\mathcal{S}|{=}200$ setting. Case~1: an immersive scrolling web page on space exploration history with illustrations. Case~2: a calculus derivatives teaching package comprising an animation video and a companion PDF handout. Case~3: a quantum computing fundamentals presentation with generated illustrations. In all cases, \emph{AgentSkillOS} produces substantially higher-quality artifacts by effectively selecting and composing relevant skills, whereas the vanilla baseline, lacking structured skill discovery and orchestration, yields visually plain or text-heavy outputs.}
\label{fig:case_study}
\end{figure*}

\paragraph{\emph{AgentSkillOS} produces more professional and usable artifacts than skill-free generation.}
Figure~\ref{fig:case_study} presents three representative tasks from the $|\mathcal{S}|{=}200$ setting, comparing the vanilla Claude Code baseline (left) with the Quality-First variant of \emph{AgentSkillOS} (right).
Case~1 requires building an immersive scrolling web page for a space exploration exhibit with chapter illustrations. The vanilla baseline produces a functional but visually plain page with small, low-fidelity images, whereas \emph{AgentSkillOS} leverages image-generation and frontend-design skills to deliver full-screen sections with high-quality illustrations and layered scrolling effects.
Case~2 asks for a calculus teaching package comprising a derivative animation and a companion handout. The vanilla baseline generates basic matplotlib plots without smooth transitions, while \emph{AgentSkillOS} invokes an animation skill to produce a polished Manim-rendered video with dynamic secant-to-tangent progression and clean mathematical annotations, alongside a well-structured PDF handout.
Case~3 requires generating a presentation on quantum computing fundamentals. The vanilla baseline produces text-heavy slides with minimal visual design, whereas \emph{AgentSkillOS} composes illustration-generation and presentation skills to produce professionally designed slides with Bloch sphere visualizations, structured layouts, and consistent branding.
Across all three cases, the quality gap stems not from differences in backbone model ability but from whether the agent successfully discovers and invokes the appropriate skills, corroborating the quantitative findings above.

\section{Related Work}
\subsection{Tool-Augmented and Multi-Skill Agents}

The integration of external tools into LLM-based agents has become a prevalent paradigm for extending model capabilities~\citep{schick2023toolformer, hao2023toolkengpt, qin2024tool}. Within such systems, LLMs are provisioned with access to APIs, functions, and execution environments~\citep{yao2022react}, allowing them to perform computation~\citep{gao2023pal}, retrieve information~\citep{lewis2020retrieval}, or interact with external software during inference~\citep{shen2023hugginggpt, wang2024executable}. 
As tool ecosystems grow in scale and diversity, \textit{skill} abstractions have been introduced as higher-level primitives that package procedural knowledge, executable resources, and structured instructions into self-contained, composable modules \citep{anthropic_agent_skills_overview}. Unlike individual tools, which typically expose a single function or API endpoint, skills encapsulate composite workflows or domain-specific procedures, enabling modular encapsulation and scalable reuse across users and platforms.
Despite these advances, new challenges emerge when skill repositories scale into large, decentralized ecosystems. First, human users struggle to identify the most relevant skills from repositories containing tens of thousands of entries, especially when skills are contributed by decentralized developers with heterogeneous naming conventions, overlapping functionalities, and varying quality levels. Without a structured overview of the ecosystem, users often lack visibility into capability coverage and functional boundaries, making targeted skill selection both time-consuming and unreliable. Moreover, it is difficult for LLM agents to effectively compose and coordinate diverse skills to handle complex, multi-step tasks, as such composition requires reasoning about inter-skill dependencies, data flow, execution order, and potential conflicts across heterogeneous tools.
To this end, we present \emph{AgentSkillOS}, the first framework for the structural organization and orchestration of open, evolving skill ecosystems at scale.
Concretely, \emph{AgentSkillOS} introduces a capability-tree-based organization mechanism for hierarchical skill discovery, together with DAG-based multi-skill orchestration that enables structured composition and execution across heterogeneous skills.

\subsection{Agent Benchmarks for Complex Tasks}
The rapid advancement of LLM-based agents has spurred the development of numerous benchmarks to evaluate their capabilities across diverse scenarios. Domain-specific benchmarks provide rigorous evaluation within individual domains such as code repair~\citep{jimenez2024swe}, web navigation~\citep{zhou2024webarena}, OS-level interaction~\citep{xie2024osworld}, and scientific computing~\citep{chen2025scienceagentbench}. Tool-use benchmarks such as ToolLLM~\citep{qintoolllm}, API-Bank~\citep{li2023api}, Gorilla~\citep{patil2024gorilla}, and TaskBench~\citep{shen2024taskbench} assess agents' ability to select and invoke external APIs with increasing scale and compositional complexity. Comprehensive benchmarks such as AgentBench~\citep{liuagentbench}, GAIA~\citep{mialon2023gaia}, and TheAgentCompany~\citep{xutheagentcompany} further broaden coverage by spanning multiple environments or task types. 
Despite their breadth, these benchmarks share two structural constraints: they confine agents to predefined and finite sets of tools rather than open-ended skill spaces; and they primarily emphasize endpoint success metrics over qualitative assessment of outputs. Therefore, they provide limited insight into agents' ability to leverage skills in more open and generative settings.
A concurrent work SkillsBench~\citep{li2026skillsbench} primarily measures the marginal performance gain from curated skills. It focuses predominantly on terminal-based tasks, limiting its ability to evaluate multi-modal or artifact-rich outputs, where the impact of skill augmentation remains underexplored.
More importantly, SkillsBench does not explicitly evaluate an agent's ability to autonomously discover, select, or compose skills across tasks. In this paper, we construct a benchmark that situates agents in an open-ended skill space spanning data analysis, visualization, document generation, and multimedia creation, requiring them to autonomously select and compose skills within each task. Beyond binary pass/fail evaluation, we adopt an LLM-based pairwise comparison protocol and aggregate the results through a Bradley-Terry model to derive unified quality scores, enabling fine-grained differentiation of agent performance on open-ended, artifact-rich tasks.

\section{Conclusions and Future Work}
Skills have proven highly effective at extending LLM-based agents with domain-specific capabilities. However, there exists a fundamental tension between the rapidly expanding skill ecosystem and the agent's inherently limited ability to discover and invoke skills, which requires an effective intermediate layer for skill management.
To address this challenge, this paper introduces \emph{AgentSkillOS}, the first principled framework that addresses this challenge through capability-tree-based skill organization, DAG-based multi-skill orchestration, and automated task execution. We further construct a benchmark of 30 artifact-rich tasks across five categories with an LLM-based pairwise evaluation protocol for fine-grained quality assessment. Through comprehensive experiments across different ecosystem scales, we demonstrate two key findings: (i) tree-based retrieval effectively approximates oracle skill selection, enabling agents to discover the most relevant skills from large-scale ecosystems; and (ii) DAG-based skill orchestration substantially outperforms native flat skill invocation even when given the identical skill set, unlocking significantly more potential from the skill ecosystem through structured composition.

We believe that two important directions remain for future work. First, our current framework assumes that skills are already collected and available; automated skill collection, including discovery of new skills from open sources, quality assessment, and continuous integration into the ecosystem, is a natural next step. Second, since skills are inherently readable artifacts, they naturally lend themselves to self-evolution, where agents automatically refine skill instructions, fix failure modes, and generate higher-quality skill variants based on execution feedback.

\bibliography{reference}
\bibliographystyle{tmlr}

\end{document}